\documentclass[manuscript]{acmart}

\AtBeginDocument{%
  }

\setcopyright{rightsretained}
\acmConference{AutomationXP26 Workshop of the 2026 CHI Conference on Human Factors in Computing Systems, April 14, 2026, Barcelona, Spain}{}{}
\acmDOI{}
\acmISBN{}


\usepackage[dvipsnames]{xcolor}
\usepackage[acronym]{glossaries}
\usepackage{wrapfig}
\usepackage{tabularx}  
\usepackage{booktabs}
\usepackage{etoolbox}
\makeglossaries
\definecolor{LightBlue}{HTML}{729CEC}
\definecolor{DarkBlue}{HTML}{3A5AA4}
\definecolor{ProblemOrange}{HTML}{fba13b}
\definecolor{OrchestratorRed}{HTML}{EC6E69}
\newacronym{llm}{LLM}{large language model}
\newacronym{mapek}{MAPE-K}{Monitor-Analyze-Plan-Execute over shared Knowledge}
\newacronym{srk}{SRK}{Skill-Rule-Knowledge}
\newacronym{react}{REACT}{Reasoning and Acting}
\newacronym{ai}{AI}{artificial intelligence}
\newacronym{rag}{RAG}{Retrieval-Augmented Generation}
\newacronym{cws}{CWS}{company world state}
\newacronym{sop}{SOP}{standard operating procedure}
\newacronym{kg}{KG}{knowledge graph}
\newacronym{roi}{ROI}{return on investment}

\begin{document}

\title[Do We Have the Knowledge We Need? Rethinking Human-AI Decision-Making in Corporations]{Do We Have the Knowledge We Need?\newline Rethinking Human-AI Decision-Making in Corporations}

\author{Anne Marx}
\email{anne.marx@ai.ethz.ch}
\affiliation{
  \institution{Department of Computer Science \& ETH AI Center, ETH Zurich}
  \country{Switzerland}
}

\author{Ricardo Maia Avelino}
\email{mricardo@ethz.ch}
\affiliation{%
  \institution{Department of Computer Science \& Architecture, ETH Zurich}
  \country{Switzerland}
}

\author{Torbj{\o}rn Netland}
\email{tnetland@ethz.ch}
\affiliation{%
  \institution{Department of Management, Technology, and Economics, ETH Zurich}
  \country{Switzerland}
}

\author{Mennatallah El-Assady}
\email{menna.elassady@ai.ethz.ch}
\affiliation{%
  \institution{Department of Computer Science, ETH Zurich}
  \country{Switzerland}
}

\renewcommand{\shortauthors}{Marx et al.}
\authorsaddresses{}

\begin{abstract}

\textit{Abstract.} Organizational knowledge is fragmented across a variety of software systems, tacit expertise, and manual documents that have traditionally been designed for human consumption. As AI systems are increasingly deployed and granted decision-making roles, they require access to this knowledge. This raises two questions: how should organizations store and maintain knowledge so that it remains accessible to both humans and future AI systems, and how should agency be allocated between humans and AI across tasks with different risks and levels of uncertainty? In this position paper, we describe how organizational knowledge evolves and contribute a framework that maps task attributes and knowledge availability to recommended agency allocations and control mechanisms. We illustrate the applicability of the framework on two different manufacturing tasks: a routine operation (visual quality inspection) and a one-off strategic decision (factory location), and conclude with opportunities for future research.

\end{abstract}


\begin{teaserfigure}
    \centering
    \includegraphics[width=\linewidth]{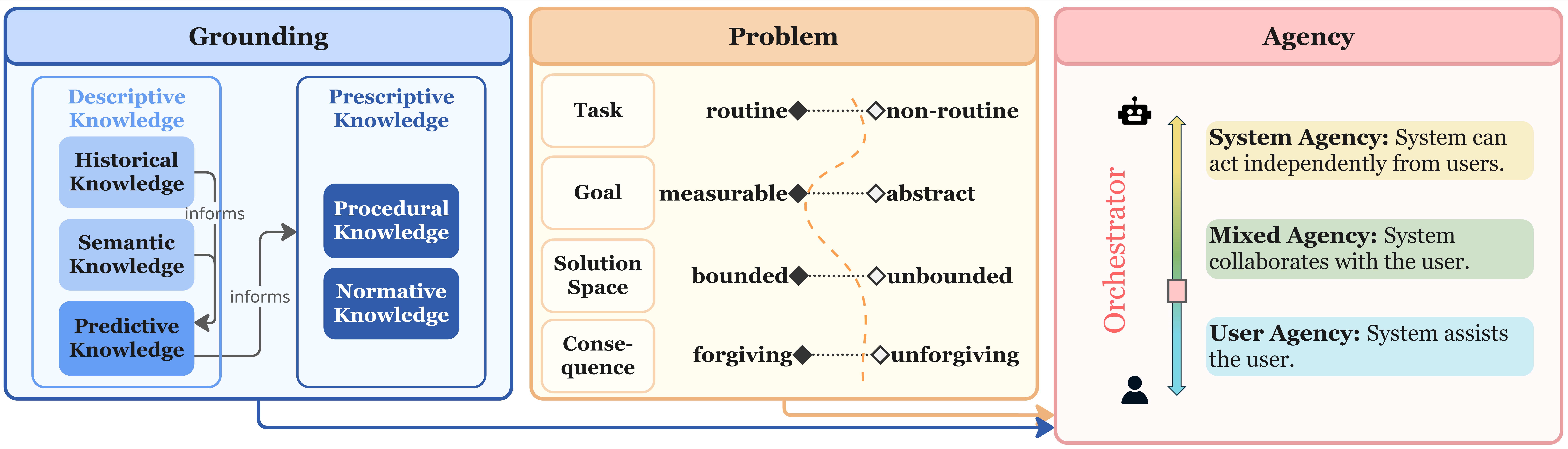}
    \Description[Teaser describing our framework for mixed agency]{In human-AI collaboration, an orchestrator can dynamically distribute agency between user and system. The agency scale has three anchors: User Agency, where the system assists the user; Mixed Agency, where the system collaborates with the user; and System Agency, where the system can act independently of users. The orchestrator makes this decision based on Grounding and Problem. Grounding refers to the basis on which we make decisions. It distinguishes between descriptive knowledge, which includes historical, semantic, and predictive knowledge, and prescriptive knowledge, which includes procedural and normative knowledge. Predictive knowledge is informed by historical and semantic knowledge and, in turn, informs prescriptive knowledge. The problem can be characterized by four attributes: The task can be routine or non-routine. The goal can be measurable or abstract. The solution space can be bounded or unbounded. The consequences can be forgiving or unforgiving. All these influence the orchestrator's decision to distribute the Agency. Keywords: knowledge theory, }
    \caption{Our mixed agency model for decision-making in manufacturing consists of a) grounding on available knowledge, b) problem characteristics that impact c) the level of agency.}
    \label{fig:teaser}
\end{teaserfigure}

\maketitle
\pagestyle{plain}
\thispagestyle{plain}
\section{Introduction}

The rapid integration of \gls{ai} into companies and their increasing autonomy levels has shifted the bottleneck for safe and effective deployment from computational power to \textit{knowledge grounding}. Current \gls{ai} behavior can be unpredictable in novel out-of-distribution situations and erroneous decisions may have considerable negative physical or economic consequences. Furthermore, increased automation creates the risk of \textit{uninvention}, which is the loss of human skill and domain knowledge when processes are automated without being fully understood or semantically represented. This reignites the long-standing challenge of selecting the level of system autonomy and automation \cite{parasuraman2000model}, now rephrased as the distribution of agency between human and AI when deciding and acting.

In companies, we can define actions as chains of making decisions and executing them, which alter the \textbf{\gls{cws}}: \textit{The current snapshot of everything that the company depends on and that the company can change}, including but not limited to physical assets, digital systems, customer relations, choice of suppliers, and its people and skills.
For a more granular analysis of agency distribution in companies, we describe \textit{world state-altering actions} with the \textbf{TDE-loop}: A formalized model describing the stages of \textit{trigger}, \textit{decision}, \textit{execution}, and optional \textit{control}.

Prior work often treats autonomy and agency interchangeably \cite{feng2025levels}, focuses on liability \cite{soder2024levels} or on executive power \cite{feng2025levels}. We explicitly separate task complexity from autonomy, as even very simple, well-defined tasks can be fully automated, yielding high cognitive autonomy and executive power. Building on human-AI collaboration taxonomies \cite{holter_deconstructing_2024, holter_2026}, we propose a dynamic allocation of agency within the TDE-loop by a human or \gls{ai} \textit{orchestrator}, which is informed by knowledge \textbf{grounding} and the specific characteristics of the \textbf{problem} at hand (see Figure \ref{fig:teaser}). Our contribution consists of (1) A dynamic model for agency: A theoretical framework that links three levels of agency (\textit{user, mixed, and system}) to a taxonomy of corporate knowledge and problem attributes, and (2) a taxonomy of grounding: We distinguish between \textit{descriptive knowledge} (historical, semantic, predictive) and \textit{prescriptive knowledge} (procedural, normative), providing a roadmap for what industrial knowledge must be captured and used for increased AI autonomy and robustness.

\section{\color{OrchestratorRed}{Agency}\color{black}{ over the TDE-Loop for World State-Altering Actions}}
Previous work formalizes actions in various domains. Human-centric domains \cite{laird2019soar, rasmussen2012skills} emphasize decision-making grounded in knowledge, skills, and context, while system-centric frameworks often aim to define an abstract, generalized procedure: The \gls{mapek}-loop is a widely recognized engineering framework for autonomic and self-adaptive systems \cite{kephart2003vision}. \gls{react} describes \gls{llm} agents \cite{yao2022react} by distinguishing thinking (reasoning) and acting, showing similarities to Kahneman's \cite{kahneman2011thinking} slow and fast thinking models. In robotics, scholars talk about the Sense-Plan-Act loop \cite{srivastava2019sense}. Both human-centric and system-centric frameworks lack mixed-agency scenarios. Generalizing these frameworks to human-only, mixed human-\gls{ai} and \gls{ai}-only decision-making for world state-altering actions, we propose the following TDE-loop with the stages: $ \textbf{T}rigger \rightarrow \textbf{D}ecision (\rightarrow Control) \rightarrow \textbf{E}xecution (\rightarrow Control)$.

Every action has a $trigger$ that initiates the loop. A $decision$ follows, which can itself be iterative and trigger further $decisions$ or TDE-loops. Afterwards, the actor $executes$, which directly alters the \gls{cws}. Observations from the $execution$ stage itself can trigger further TDE-loops. Both $decision$ and $execution$ can be $controlled$, establishing boundaries and safeguards, especially in high-stakes settings. Agency can be split between human or system \textit{users} and AI \textit{systems} across all stages. We speak of AI \textbf{system agency} when the AI primarily bears responsibility for the TDE-loop and mostly executes tasks autonomously. In \textbf{mixed agency} settings, both the user and the system collaborate, whereas in \textbf{user agency}, the user owns the TDE-loop, with the AI system at most assisting the user on demand.

\section{\color{DarkBlue}{Grounding}\color{black}{ -- Descriptive and Prescriptive Knowledge}}
The availability of different types of knowledge shapes how we can make decisions and impacts agency. We distinguish grounding into \textit{descriptive} (historical, semantic, predictive) and \textit{prescriptive} (procedural, normative) knowledge, drawing on different organizational and cognitive views of knowledge, including Nonaka's SECI Model \cite{farnese2019managing}, Tulving's episodic and semantic memory \cite{tulving1972episodic}, and Ackoff's DIKW hierarchy \cite{tuomi1999data} while adapting them to \gls{ai}-enabled work (see Appdx. \ref{appendix:grounding}).

\vspace{4pt}\textbf{\color{LightBlue}{Descriptive knowledge}} describes the world and can be sub-categorized as follows:
\begin{itemize}
    \item ``What happened?'' \textit{Historical knowledge} describes any historical data that was extracted from perceiving the \gls{cws}, such as human observations, data generated from machines, or general state-specific information.
    \item ``What is?'' \textit{Semantic knowledge} describes (time- and stateless) facts about the company's world, such as definitions, attributes (e.g., physical properties) or relations.
    \item ``What will (most likely) happen?'' \textit{Predictive knowledge} enables the recognition of similar historic situations or logical deductions based on semantics to derive predictions of future \glspl{cws}.
\end{itemize}
Historical and semantic knowledge are directly related to Tulving's episodic and semantic memory \cite{tulving1972episodic}. Once historical data and/or semantic knowledge is collected, it may be possible to predict future \glspl{cws}, for example, by training \gls{ai} models on historical data to perform predictions and thus generate predictive knowledge.

\vspace{4pt} \textbf{\color{DarkBlue}{Prescriptive knowledge}} is derived from predictive knowledge, as understanding or estimating cause-and-effect relationships leads to formulating best practices, manuals, or rules. Prescriptive knowledge is composed of:
\begin{itemize}
    \item ``What am I allowed and not allowed to do?'' \textit{Normative knowledge} describes rules and constraints for how to act, such as safety rules, company policies, usage norms, etc.
    \item ``How should or can I do that?'' \textit{Procedural knowledge} describes the step-by-step sequence of how to act for a predefined task, often resulting in best practice guides and manuals.
\end{itemize}

\section{\color{ProblemOrange}{Problem Attributes}\color{black}{ -- Regularity, Consequences, Goals and Solution Space}}
\label{subsection:dimensions}
While \textit{grounding} provides the foundation for decision-making, we also need characteristics of the \textit{problem} to determine how we can make the decision and who receives agency. We argue that the key problem attributes for the orchestrator to decide who should have authority over the TDE-loop include whether the situation is \textit{routine}, the severity of the \textit{consequences}, the clarity of the \textit{goals}, and the boundedness of the \textit{solution-space}.

The problem's regularity is related to the availability of knowledge. The human-centric \gls{srk} framework from Rasmussen \cite{rasmussen2012skills} differentiates decision-making for human operators in (1) skill-based behavior for routine tasks, when not much cognitive effort is required by the operator (implicit procedural and predictive knowledge), (2) rule-based behavior for following explicit guidelines and procedures (explicit prescriptive knowledge), and (3) knowledge-based behavior for goal-controlled tasks in unfamiliar situations (normative and descriptive knowledge). Rasmussen also shows that procedural knowledge tasks lead to fewer errors than knowledge-based tasks \cite{rasmussen2012skills}.
However, procedural knowledge may be hard to obtain, e.g, if the task occurs rarely and mostly under novel circumstances. In these situations, normative knowledge and descriptive knowledge become increasingly important signals. 

Perrow et al. \cite{perrow2011normal} distinguish between loose coupling (forgiving) and tight coupling (rather unforgiving consequences) in systems. In manufacturing, a decision made in a tightly coupled system, such as a chemical plant, can propagate errors, potentially catastrophically. Generally, high-stakes actions in these situations follow standardized procedures to reduce errors and are subjected to additional $controls$. The possible consequences, coupled with the novelty of the situation, are thus a crucial signal for the orchestrator to decide who should have authority over the TDE-loop.

Similarly, decision-making can follow clear, measurable goals, such as finding the cheapest vendor, or be more abstract and ambiguous, such as improving workers' health, where human intuition can be useful for clarifying the task. 

In addition, the solution space may be wider or narrower. Rittel et al. \cite{rittel1973dilemmas} describe the ``Wicked Problem'' as having no clear final solution and thus no stopping rule. While AI-systems can navigate wider solution spaces than humans, they struggle to act when no measurable goals have been defined. This trade-off is then critical for distributing agency as illustrated in the following case studies.

\section{Case Studies}
We apply our framework to two problems recently observed by the authors in manufacturing companies.

\textit{\textbf{Visual Quality Control.}} An incoming car part (\textit{trigger}) is tested for visual defects in the paint job (\textit{decision}) and sorted out (\textit{execution}) as faulty or proceed to the next manufacturing stage.
\textbf{\color{DarkBlue}{Grounding}}: The company has collected thousands of historical examples and predictive annotations. The \textbf{\color{ProblemOrange}{problem}} has the following characteristics: \textit{routine} problem, clear and \textit{measurable} goal, \textit{bounded} solution space (accept/reject), and \textit{forgiving} consequences. \begin{wrapfigure}{r}{0.4\textwidth}
    \setlength{\columnsep}{0pt}
    \centering
    \vspace{-15pt}
    \includegraphics[width=0.95\linewidth]{system_agency.jpg}
    \Description{In this case study, we have system agency over all stages of the TDE-loop for the standard case. Tags: autonomy, system agency}
    \vspace{-20pt}
\end{wrapfigure}\textbf{\color{OrchestratorRed}{Agency}}: An \gls{ai} model trains on the annotated data and, in the process, develops \textit{implicit procedural knowledge}. The orchestrator chooses system agency in the standard setting. If a novel car part is introduced, mixed agency is used, with the consequences being forgiving: System agency is used in the \textit{decision} and \textit{execution}, while a human \textit{controls}. If the system proves unreliable on the novel part, human feedback can be used to update the model's insufficient procedural knowledge. Or, explicit semantic knowledge, such as ``scratches or missing paint are types of errors,'' enables the use of specialized \gls{ai} models that detect scratches and color changes, e.g., from other domains, or general vision language models, thus making explicit decisions for more robust system actions in these non-routine situations.

\textit{\textbf{Opening a New Factory.}} In a novel geopolitical situation (\textit{trigger)}, a company must choose a new factory location (\textit{decision}) and initiate its construction (\textit{execution}). 
\textbf{\color{DarkBlue}{Grounding}}: The knowledge is dispersed into various types of descriptive knowledge, such as historical national statistics, logistics models, geopolitical developments, taxes, and expected branding impact. The \textbf{\color{ProblemOrange}{problem}} has the following characteristics: \textit{Non-routine} problem, \textit{measurable} and clear goal, \textit{big} solution space, and \textit{unforgiving} consequences. The \textbf{\color{OrchestratorRed}{agency}} can be mixed if procedural knowledge (such as a playbook for site selection with clear criteria) or semantic knowledge (such as ontologies defining logistical resilience or political stability) is externalized and accessible to the \gls{ai}. Then, the system could execute smaller tasks such as screening locations, applying externalized logic to filter location candidates, or challenging the user. \begin{wrapfigure}[2]{r}{0.4\textwidth}
    \setlength{\columnsep}{0pt}
    \centering
    \vspace{-15pt}
    \includegraphics[width=0.95\linewidth]{mixed_agency.jpg}
    \Description{In this case study, we have mixed agency over the TDE-loop: User agency in the trigger and control stages, and mixed agency for the decision and execution stages. Tags: human-AI collaboration, collaborative decision-making}
    \vspace{-20pt}
\end{wrapfigure}Full system agency is unlikely due to the lack of logical reasoning in \gls{ai}, and $control$ must remain with the user. Furthermore, managing stakeholders in this process requires applying nuanced social norms, which so far is reserved for humans.

\section{Opportunities: Bridging the Agency-Grounding Gap}
\label{sec:opportunities}

When available data perfectly overlaps with the problem space, AI achieves high autonomy. However, the industry currently over-optimizes for \textit{historical} and \textit{procedural} knowledge, neglecting the \textit{semantic} and \textit{normative} layers. We identify three critical frontiers for research.

\textbf{\textit{Resilience-First Design.}} 
Gorman \cite{gorman2002types} warns that the failure to preserve tacit knowledge leads to ``uninvention,'' while Ganuthula \cite{ganuthula2024paradox} warns about ``unlearning'', indicating that \gls{ai} usage risks human operators losing the critical skills required to intervene in non-routine failures. We could expect a shift from Efficiency-First to Intentional Inefficiency: systems should occasionally return agency to humans—even for fully automatable tasks—to maintain ``skill resilience'' and increase robustness in novel situations, and, in addition, capture semantic knowledge for on-demand access. The orchestrator must also optimize for maintaining the human-in-the-loop as a robust fallback system.

\textbf{\textit{Mining the Reasoning Layer.}} 
Current documentation is often designed for human consumption, preventing \gls{ai} from tracing the ``why'' behind a decision. This opens a vast research opportunity: How can we automatically mine semantic, normative, and procedural knowledge from unstructured legacy data, multi-modal observations, or from human interaction? Emerging work in knowledge extraction \cite{de2026procedural} demonstrates the potential of using vision language models to convert static manuals into \glspl{kg}. By tracing the provenance of \textit{procedural} knowledge back to \textit{normative}, \textit{semantic}, or \textit{historic} information, we could further enable \gls{ai} to act more reliably in novel situations where statistical correlations fail. We further ask: What is the optimal knowledge storage for both human and \gls{ai} accessibility? How will this change \gls{ai} reliability in novel situations and future human-\gls{ai} collaboration?

\textbf{\textit{Agency as a Dynamic Variable.}} 
One way to capture semantics is to enable future systems to move beyond static automation toward dynamic agency negotiation. The system self-assesses its grounding by checking if the current problem attributes map to its available knowledge before requesting or accepting authority. If the system detects a ``grounding gap'' (e.g., in our visual quality control case, a novel part with unknown semantic properties appears), it should proactively degrade its agency level and request specific human input to fill the knowledge gap.

\section{Conclusion and Future Outlook}
\label{sec:conclusion}

We argued that the bottleneck to autonomy and effective human-AI collaboration in companies lies in \textit{knowledge grounding}. We introduced the \textit{TDE-loop} to formalize actions that alter the \textit{company world state} and provide a framework for dynamically allocating agency based on the alignment of knowledge types and problem attributes.

Achieving reliable mixed agency requires a fundamental shift in how we view corporate knowledge. We conclude with four provocations for the future of human-AI collaboration in companies:
\begin{itemize}
    \item \textbf{Explicit Knowledge as the New Digital Gold:} When high-quality training data is scarce, and models rely on context, the competitive advantage will shift to proprietary \textit{reasoning layers}. Specialized and industry-specific knowledge storages with high accessibility may become as valuable as the \gls{ai} models themselves, leading to a marketplace for ``corporate wisdom'' rather than just pretrained transformers.
    \item \textbf{Closing the Gap to Reality:} Externalized knowledge can vastly differ from values, processes, and actions taken in reality. These discrepancies will need to be iteratively checked and overcome through feedback loops, e.g., through provenance modeling or interactive sessions. Resolving conflicting realities and norms can open up further research opportunities.
    \item \textbf{Graph-Optimized AI Architectures:} Backed by the resurgence of neuro-symbolic AI \cite{delong2024neurosymbolic}, we foresee the rise of architectures co-designed with a graph-based knowledge storage to provide native reasoning. These systems will not only retrieve context but also follow logic more reliably, enabling them to navigate irregular problems.
    \item \textbf{The Orchestrator as a Mediator:} The orchestrator of the future will serve as a knowledge farmer and bi-directional translator, such as converting human \textit{normative} intent into explicit machine-executable constraints, and translating machine \textit{predictive} confidence into human-understandable agency requests.
\end{itemize}
By treating knowledge as a dynamic infrastructure accessible to both humans and AI, we can transition from rigid automation to resilient, mixed agency environments that navigate the unpredictable complexities of the modern world.

\begin{acks}
This research was primarily supported by the ETH AI Center through an ETH AI Center doctoral fellowship to Anne Marx. Additional support was provided by Design++ and Halter Group AG through a postdoctoral fellowship to Ricardo Maia Avelino, and by the Swiss National Science Foundation (SNSF) under grant number 10003068.
\end{acks}

\bibliographystyle{ACM-Reference-Format}
\bibliography{sample-base}
\printglossaries
\appendix
\section{Theoretical Foundations of Grounding}
\label{appendix:grounding}

Our taxonomy of knowledge grounding synthesizes concepts from cognitive science and organizational theory, adapted for human-AI collaboration. Nonaka's famous SECI model \cite{farnese2019managing} focuses on the cycle of converting tacit to explicit (codified) knowledge in organizations. Parts of our descriptive knowledge mirror Tulving's cognitive distinction between episodic memory (time-stamped events) and semantic memory (general facts) \cite{tulving1972episodic}. Both works do not account for \gls{ai} systems and how actionable knowledge develops. Finally, Ackoff's influential DIKW hierarchy focuses on the distinction of data, information, knowledge, and wisdom, which has in the past been criticized, as it is disputable whether knowledge or data comes first \cite{tuomi1999data}. Although we agree on the hypothesis that causal relationships enable forecasting, our framework does not distinguish among mere data, information, and knowledge.

\section{Comparison of Action Loops}
\label{appendix:loops}

The TDE-loop formalizes world state-altering actions in a way that allows for dynamic agency allocation, differing from existing system-centric loops:

\begin{itemize}
    \item \textbf{vs. MAPE-K \cite{kephart2003vision}:} While MAPE-K is an engineering loop for self-adaptive systems, it misses to integrate humans. The TDE further explicitly introduces an \textit{orchestrator} to negotiate authority between human and AI at each stage.
    \item \textbf{vs. Sense-Plan-Act \cite{srivastava2019sense}:} TDE accounts for ``triggers'' external to the system's sensors and inserts a ``control'' stage, which is essential for the \textit{unforgiving consequences} common in manufacturing.
    \item \textbf{vs. ReAct \cite{yao2022react}:} TDE also separates the cognitive \textit{decision} from the physical/digital \textit{execution} while further allowing to verify reasoning before the world state is altered.
\end{itemize}

\section{The Orchestrator Logic for Agency Allocation}
\label{appendix:orchestrator}

The following logic describes how agency can be assigned by the orchestrator:

\begin{itemize}
    \item \textbf{System Agency:} Appropriate when grounding is high (historical + predictive), the task is routine, and consequences are forgiving.
    \item \textbf{Mixed Agency:} Required for non-routine tasks with possibly high consequences in collaborative settings. Here, the system may suggest \textit{decisions} but the user usually directly retains authority over \textit{execution}, or indirectly through \textit{control} agency. In other settings, the system can also \textit{control} user actions for increased efficiency or safety.
    \item \textbf{User Agency:} Mandatory for abstract goals and novel situations where procedural, normative, or semantic logic is not machine-accessible.
\end{itemize}

\section{Knowledge Graphs as Semantic Infrastructure}
\label{appendix:kg}

We advocate for \glspl{kg} as a reasoning layer.
\begin{enumerate}
    \item \textbf{Generalization:} KGs allow AI to reason about novel parts by following semantic links (e.g., ``Part X is a subset of Component Y'').
    \item \textbf{Skill Retention:} By externalizing expert logic into graph nodes and relations, organizations mitigate the risk of ``uninvention'' as domain experts retire and automation levels increase. Knowledge can then be dynamically accessed on-demand through a human-interpretable interface.
    \item \textbf{Explainability:} In mixed-agency settings where AI uses KGs to base it's decisions, the human can trace the arguments and reasoning back to the KG.
\end{enumerate}
\end{document}